\documentclass[11pt]{article}

\usepackage{times}
\usepackage{latexsym}
\usepackage{graphicx}

\usepackage[T1]{fontenc}

\usepackage[utf8,utf8x]{inputenc}

\usepackage{microtype}

\usepackage{inconsolata}

\usepackage[]{acl}

\usepackage{subcaption}
\usepackage{paralist}

\setdefaultleftmargin{0em}{}{}{}{}{}

\title{Knesset-DictaBERT:\\ A Hebrew Language Model for Parliamentary Proceedings}

\author{Gili Goldin \\
  University of Haifa \\
  \texttt{gili.sommer@gmail.com} \\\And
  Shuly Wintner  \\
  University of Haifa \\
  \texttt{shuly@cs.haifa.ac.il} \\}

\begin{document}
\maketitle
\begin{abstract}
We present Knesset-DictaBERT, a large Hebrew language model fine-tuned on the Knesset Corpus, which comprises Israeli parliamentary proceedings. The model is based on the DictaBERT architecture and demonstrates significant improvements in understanding parliamentary language according to the MLM task. We provide a detailed evaluation of the model's performance, showing improvements in perplexity and accuracy over the baseline DictaBERT model.
\end{abstract}

\section{Introduction}
The field of natural language processing (NLP) has seen remarkable advancements in recent years, driven by the development of large language models. These models have significantly enhanced the ability to understand and generate human language. However, much of the effort of creating and training NLP models has been focused on English, while fewer NLP models are available for lower-resource languages such as Hebrew. We present one such model here.

Parliamentary proceedings are a valuable source of information for understanding political discourse, legislative processes, and more. Analyzing these texts requires models that can accurately capture the nuances of the language used in such settings. Despite the importance of this task, there has been a lack of specialized models trained on parliamentary corpora in Hebrew.
To address this gap, we fine-tuned the pre-trained Dicta-BERT model \citep{shmidman2023dictabert} on the Knesset Corpus \citep{Goldin2024TheKC}, a dataset of Israeli parliamentary proceedings. DictaBERT is a state-of-the-art Hebrew language model based on the BERT architecture \citep{devlin-etal-2019-bert}, which was trained on a diverse set of Hebrew texts.
Knesset-DictaBERT is the resulting fine-tuned model: it is specifically tailored for Hebrew parliamentary text. In this paper we describe the training process of this model, provide a detailed evaluation of its performance, and demonstrate its superiority over the baseline DictaBERT model on the Knesset data.
We believe that Knesset-DictaBERT will be a valuable resource for researchers and other users working on Hebrew language processing and political text analysis.

\section{Methodology}
\label{sec:Methodology}
We fine-tuned the pre-trained DictaBERT model, a state of the art language model for Hebrew, specifically for the masked language modeling (MLM) task, on the full Knesset Corpus dataset \citep{Goldin2024TheKC}. The model was initialized using the pre-trained weights from the \href{https://huggingface.co/dicta-il/dictabert}{dicta-il/dictabert} checkpoint available on Hugging Face.
\subsection{Dataset and Data preprocessing}
We used the Knesset Corpus, which is is a large Hebrew dataset of Israeli parliamentary proceedings, as the dataset for fine-tuning DictaBERT. The corpus contains over 32M sentences, over 384M tokens and comprises texts from both plenary and committee protocols. 

The corpus was preprocessed to create text shards for efficient loading and processing. We split the dataset into training, validation, and test sets with ratios of 80\%, 10\%, and 10\%, respectively.

First, we tokenized the input text using the \href{https://huggingface.co/docs/transformers/v4.15.0/en/model_doc/auto#transformers.AutoTokenizer}{AutoTokenizer} from Hugging Face's transformers library. Each text sample was tokenized into sequences of tokens, with each token represented by its corresponding ID from the tokenizer's vocabulary.
To prepare the data for the MLM training, we grouped the tokenized texts into fixed length chunks of 256 tokens. The last chunk was padded to ensure consistent input sizes for the model. We used the \href{https://huggingface.co/docs/transformers/v4.42.0/en/main_classes/data_collator#transformers.DataCollatorForLanguageModeling}{DataCollatorForLanguageModeling} from the transformers library to dynamically mask tokens during training, with a masking probability of 15\%.
\subsection{Training procedure and hyperparameters}\label{Training}
We trained the model on a \href{https://slurm.schedmd.com/documentation.html}{SLURM} (Simple Linux Utility for Resource Management) environment.
We utilized a distributed training setup with the \href{https://developer.nvidia.com/nccl}{NCCL} backend to leverage multiple GPUs, ensuring efficient training and gradient synchronization.

We used the following training configuration: We used a per-device batch size of~32. In order to effectively double the batch size without increasing memory usage, we set the gradient accumulation steps to~4. A learning rate of $1e^{-4}$ was chosen for the AdamW optimizer and a weight decay of~0.01 was applied to regularize the model. The model was trained for~2 full epochs.
We enabled `fp16' (mixed precision) to speed up training and reduce memory usage. Mixed precision training uses 16-bit floating-point numbers instead of the standard 32-bit, which can significantly improve computational efficiency and decrease memory usage \cite{DBLP:journals/corr/abs-1710-03740}.
Periodic evaluations were conducted on the validation set, and the best-performing model checkpoint was identified based on the validation loss.
Final evaluations were performed on the test set to assess the model's performance. The model is available on Hugging Face hub at \href{https://huggingface.co/GiliGold/Knesset-DictaBERT}{Knesset-DictaBERT}.

\section{Experiments and Results}
The fine-tuned Knesset-DictaBERT model was evaluated on the test set, which contained about~3.2 million sentences (about 38 million tokens), using perplexity as the primary metric. The model achieved a perplexity of~6.60, significantly outperforming the original DictaBERT model, which showed a perplexity of~22.87.
Evidently, the Knesset-DictaBERT model reflects the language of Knesset proceedings much better than the baseline DictaBERT model.

We also evaluated the fine-tuned Knesset-DictaBERT model for its accuracy in predicting masked tokens on a subset of the test set containing approximately~300K sentences (about~3.5 million tokens).  Since MLM is considered a challenging task, where multiple options may be equally plausible, the evaluation focused on three accuracy metrics: top-1 accuracy, top-2 accuracy, and top-5 accuracy. These metrics measure the model's ability to correctly predict the masked token as the first, within the first two, or within the first five predictions, respectively.

The Knesset-DictaBERT model correctly identified the masked token in the top-1 prediction 52.55\% of the time, compared to the original Dicta model, which achieved a top-1 accuracy of 48.02\%. Additionally, Knesset-DictaBERT succeeded in cases where the original DictaBERT model did not a total of 52,464 times. In contrast, the original DictaBERT model succeeded where Knesset-DictaBERT did not in only 27,995 times.
Furthermore, when considering the top-2 predictions, Knesset-DictaBERT correctly identified the masked token 63.07\% of the time, whereas the original DictaBERT model had a top-2 accuracy of 58.60\%. Moreover, Knesset-DictaBERT succeeded in 19,400 instances where the original model failed to provide a correct prediction within the top-2, while the original DictaBERT model, succeeded in only 13,862 instances where Knesset-DictaBERT did not. 
On top of that, when extending the scope to the top-5 predictions, Knesset-DictaBERT demonstrated a significant improvement with a 73.59\% accuracy, while the original DictaBERT model achieved a 68.98\% accuracy. 

In all tested metrics the Knesset-DictaBERT model outperformed the original DictaBERT model, indicating a more robust performance in predicting masked tokens within parliamentary text. These results highlight the effectiveness of fine-tuning on the specific parliamentary corpus.
The results are presented in Table~\ref{tbl:accuracy_table}.

\begin{table*}[hbt]
\centering
\begin{tabular}{|l|c|c|}
\hline
\textbf{Metric} & \textbf{Knesset-DictaBERT} & \textbf{Original DictaBERT} \\
\hline
\hline

Perplexity & \textbf{6.60} & 22.87 \\
\hline
\hline

Top-1 Accuracy & \textbf{52.55\%} & 48.02\% \\
\hline
Top-2 Accuracy & \textbf{63.07\%} & 58.60\% \\
\hline
Top-5 Accuracy & \textbf{73.59\%} & 68.98\% \\
\hline
\end{tabular}
\caption{Comparison of Knesset-DictaBERT and Original DictaBERT on perplexity and accuracy metrics}
\label{tbl:accuracy_table}
\end{table*}

\section{Conclusion and Future Work}\label{conclusions}
In this work, we successfully fine-tuned the DictaBERT model on the Knesset Corpus to create Knesset-DictaBERT, a model proficient at understanding and generating parliamentary language in Hebrew. The results indicate a robust model performance, with substantial improvements over the baseline model. Future work may involve evaluation on additional Hebrew datasets to enhance the model's generalization capabilities and fine-tuning other language models on the Knesset corpus.

\section*{Limitations}
The model was fine-tuned specifically on the Knesset Corpus, which comprises parliamentary proceedings. As a result, its performance on general Hebrew text or other domains may not be as robust. However, the original DictaBERT model was trained on a variety of resources in Hebrew, which probably allows the Knesset-DictaBERT to still benefit from the diverse linguistic patterns and vocabulary present in the broader training data of the original model.

\section*{Ethical Considerations}
The Knesset Corpus may contain inherent biases, reflecting the political and social biases present in parliamentary discussions. Consequently, Knesset-DictaBERT may inherit these biases.

\bibliography{anthology,custom}

\appendix

\end{document}